\documentclass[runningheads,a4paper]{llncs}

\usepackage{amssymb}
\usepackage{subfigure}
\usepackage{graphicx}
\usepackage{bm}
\usepackage{amsmath}
\usepackage{url}
\usepackage{algorithm}
\usepackage{algorithmic}
\usepackage{multirow}
\usepackage{color}

\urldef{\mailsa}\path|{smohamad, abouchachia}@bournemouth.ac.uk|    
\newcommand{\keywords}[1]{\par\addvspace\baselineskip
\noindent\keywordname\enspace\ignorespaces#1}

\begin{document}

\mainmatter  

\title{Online Gaussian LDA for Unsupervised Pattern Mining from Utility Usage Data}

\titlerunning{Online Gaussian LDA for Pattern Mining from Utility Usage Data}

%
%
\author{Saad Mohamad \and Abdelhamid Bouchachia}
\authorrunning{S. Mohamad, C. Mansouri and A. Bouchachia}

\institute{Department of Computing,\\
Bournemouth University, \\
 Poole, UK
}

%
%

\toctitle{Lecture Notes in Computer Science}
\tocauthor{Authors' Instructions}
\maketitle

\begin{abstract}
Non-intrusive load monitoring (NILM) aims at separating a whole-home energy signal into its appliance components. Such method can be harnessed to provide various services to better manage and control energy consumption (optimal planning and saving). NILM has been traditionally approached from signal processing and electrical engineering perspectives. Recently, machine learning has started to play an important role in NILM. While most work has focused on supervised algorithms, unsupervised approaches can be more interesting and of practical use in real case scenarios. Specifically, they do not require labelled training data to be acquired from individual appliances and the algorithm can be deployed to operate on the measured aggregate data directly. In this paper, we propose a fully unsupervised NILM framework based on Bayesian hierarchical mixture models. In particular, we develop a new method based on Gaussian Latent Dirichlet Allocation (GLDA) in order to extract global components that summarise the energy signal. These components provide a representation of the consumption patterns. Designed to cope with big data, our algorithm, unlike existing NILM ones, does not focus on appliance recognition. To handle this massive data, GLDA works online. Another novelty of this work compared to the existing NILM is that the data involves different utilities (e.g, electricity, water and gas) as well as some sensors measurements. Finally, we propose different evaluation methods to analyse the results which show that our algorithm finds useful patterns.
\end{abstract}

\section{Introduction}\label{sec1}
The monitoring of human behaviour is highly relevant to many real-word domains such as safety, security, health and energy management. Research on human activity recognition (HAR) has been the key ingredient to extract pattern of human behaviour. There are three main types of HAR, sensor-based~\cite{bulling2014tutorial}, vision-based~\cite{poppe2010survey} and radio-based~\cite{wang2015review}. A common feature of these methods is that they all require equipping the living environment with  embedded devices (sensors). On the other hand, non-intrusive load monitoring (NILM) requires only single meter per house or a building that measures aggregated electrical signals at the entry point of the meter. Various techniques can then be used to disaggregate per-load power consumption from this composite signal providing energy consumption data at an appliance level granularity. In this sense, NILM focus is not extracting general human behaviour patterns but rather identifying the appliances in use. This, however, can provide insight into the energy consumption behaviour of the residents and therefore can  express users’ life style in their household. The idea of abandoning the high costs induced by various sensors entailed by traditional HAR makes NILM an attractive approach to exploit in general pattern recognition problems. On the other hand, taking the human behaviour into account can leverage the performance of NILM; thus, providing finer understanding of the resident's energy consumption behaviour. In this paper, we do not distinguish between patterns and appliances recognition. The main goal of our approach is to encode the regularities in a massive amount of energy consumption data into a reduced dimensionality representation. This is only possible by the fact that human behaves in certain pattern and not randomly. We are also lucky to have an extra large amount of real world data which makes this approach more viable. 

Since the earliest work on NILM~\cite{hart1992nonintrusive}, most NILM work has been based on signal processing and engineering  
approaches~\cite{zeifman2011nonintrusive,zoha2012non}. This can explain the fact that even with the economical attractive tools that NILM can provide for PR and HAR communities, it has not been widely exploited. Most of existing machine learning approaches to NILM adopt supervised algorithms~\cite{hart1992nonintrusive,liang2010load,kolter2010energy,srinivasan2006neural,berges2009learning,ruzzelli2010real,kelly2015neural,lai2013multi}. Such algorithms could damage the attractiveness of NILM as they require individual appliance data for training, prior to the system deployment. Hence, there is a need to install one energy meter per appliance to record appliance-specific energy consumption. This incurs extra costs and  a complex installation of sensors on every device of interest. In contrast, unsupervised algorithms can be deployed to operate directly from the measured aggregate data with no need for annotation. Hence, unsupervised algorithms are clearly more suitable for NILM. To the best of our knowledge, all existing unsupervised approaches to NILM~\cite{bonfigli2015unsupervised} concentrate on disaggregating the whole house signal into its appliances' ones. In contrast, our approach, as mentioned earlier, does not focus on identifying per-appliance signal. We instead propose a novel approach that seeks to extract human behaviour patterns from home utility usage data. These patterns could be exploited for HAR as well as energy efficiency applications. 
\begin{figure}[t]
\centering
\includegraphics[width=\textwidth,height=8.7cm]{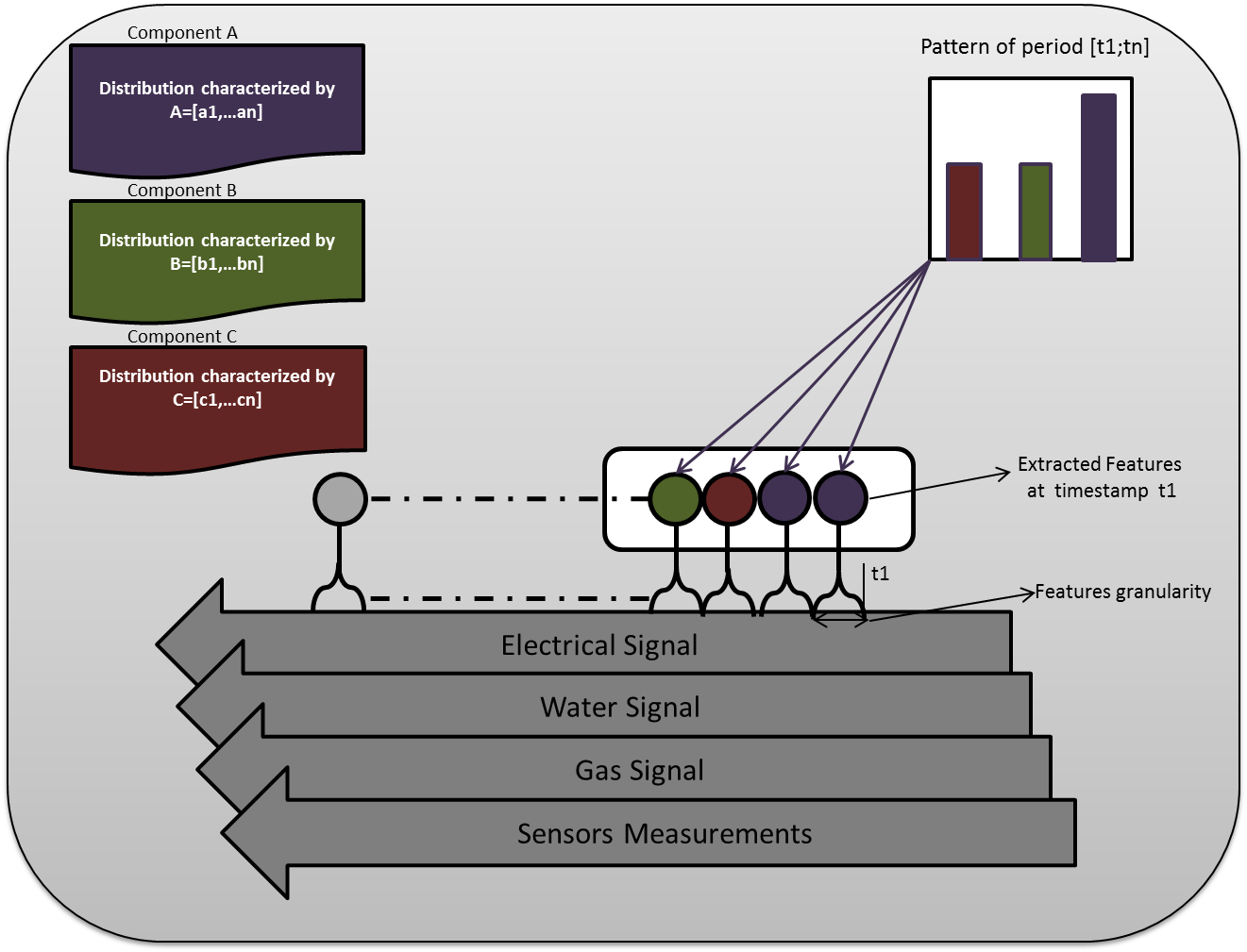}
\caption{Elements of the proposed approach}
\label{fig1}
\end{figure}

The proposed approach is based on a hierarchical Bayesian mixture model. More precisely, this model is a member of the family of graphical models proposed by~\cite{hoffman2013stochastic} where observations, global hidden variables, local hidden variables, and fixed parameters are involved. Under further assumptions in addition to the ones in~\cite{hoffman2013stochastic}, we end up with a Gaussian version of Latent Dirichlet Allocation (GDLA) where the observations are continuous and not discrete. In particular, we assume that the hidden local variables are conditionally independent; hence, the observations can be treated as a bag of words. This approach has drawn inspiration from the success that LDA has achieved in the domain of text modelling. To explain the analogy between LDA and the proposed approach, we show in Fig.~\ref{fig1} an example where three components  have been extracted from the utility usage data. Here, the components are equivalent to topics in LDA. Because the features extracted from the data are in continuous space, the components represent Gaussian distributions over the input features instead of categorical distributions over words as in LDA. A pattern is a mixture of components generating the input features over a fixed period of time. In LDA, patterns are associated with documents that can be expressed by mixture of corpus-wide topics. One can clearly notice that this bag-of-words assumption, where temporal dependency in the data is neglected, is a major simplification. However, this simplification leads to methods that are computationally efficient. Such computational efficiency is essential in our case where massive amount of data (around 4 Tb) is used to train the model. 

In this work, we demonstrate that, similar to LDA in the domain of text mining, this approach can capture significant statistical structure in a specified window of data over a period of time. This structure provides understanding of regular patterns in the human behaviour that can be harnessed to provide various services including services to improve energy efficiency. For example, understanding of the usage and energy consumption patterns could be used to predict the power
demand (load forecasting), to apply management policies and to avoid overloading the energy network. Moreover, providing consumers with information about their consumption behaviour and making them aware of abnormal consumption patterns compared to others can influence their behaviour to moderate energy consumption~\cite{fischer2008feedback}.

As already mentioned, this algorithm is going to be trained over a very huge amount of data resulting from the high sampling rate around $205$ kHz of the electricity signal which gives us an advantage compared to the data used in other research studies except for~\cite{kelly2015uk,filip2011blued,kolter2011redd}. Specifically, high sampling rate allows extraction of rich features in contrast to the limited number of features that can be extracted from low frequency data. To handle such big amount of data, online version of GLDA is developed. This can be done by defining particular distributions for the exponential family in the  class of models described in~\cite{hoffman2013stochastic}. More details can be found in Sec.~\ref{sec3}. Besides the advantage the data size offers, apart from ~\cite{makonin2013ampds,makonin2016electricity} whose sampling rate is very low, our data is the only one including water and gas usage data. Moreover, measurements provided by additional sensors are also exploited to refine the performance of the pattern recognition algorithm. More details on the data can be found in Sec.~\ref{sec4}. The diversity of the data is another motivation for adopting a pattern recognition approach rather than traditional disaggregation approach. 

The rest of the paper is organised as follows. Section~\ref{sec2} presents the related work. Section~\ref{sec3} presents the proposed approach. Section~\ref{sec4} describes the data and discusses the obtained results. Finally, Sec.~\ref{sec5} concludes the paper and hints to future work.

\section{Related Work}\label{sec2}
We divide the related work into two parts: (i) machine learning approaches to NILM and (ii) NILM data used in the literature.

As we have discussed in the introduction, most of existing NILM studies are not based on machine learning algorithms and most of  machine learning NILM algorithms are supervised ones~\cite{hart1992nonintrusive,liang2010load,kolter2010energy,srinivasan2006neural,berges2009learning,ruzzelli2010real,kelly2015neural,lai2013multi}. Such algorithms requires training on labelled data which is expensive and laborious to obtain. In fact, the practicality of NILM is stemmed from the fact that it comes with almost no setup cost. Recently, researchers have started exploring unsupervised machine learning algorithms to NILM. These methods have mainly focused on performing energy disaggregation to discern appliances from the aggregated load data directly without performing any sort of event detection. The most prominent of these methods are based on Dynamic Bayesian Network models, in particular different variants of Hidden Markov Model (HMM)~\cite{kim2011unsupervised,kolter2012approximate,johnson2013bayesian}. 

Authors in~\cite{kim2011unsupervised} proposes to use Factorial Hidden Markov Model (FHMM) and three of its variants: Factorial Hidden Semi-Markov Model (FHSMM), Conditional FHMM (CFHMM) and Conditional FHSMM (CFHSMM) to achieve energy disaggregation. The main idea is that the dynamics of the state occupancy of each appliance evolves independently and the observed aggregated signal is some joint function of all the appliances states. To better model the state occupancy duration, that is modelled with a geometric distribution  by FHMM, authors propose to use FHSMM which allows modelling the durations of the appliances states with gamma distribution. Authors also propose CFHMM to incorporate additional features, such as time of day, other sensor measurements, and dependency between appliances. To harness the advantages of FHSMM and CFHMM, authors propose a combination of the two models resulting in CFHSMM. In that work, the electricity signal was sampled at low frequency which is in contrast to our work.

Similar approach was taken in ~\cite{kolter2012approximate} where Additive Factorial Hidden Markov Model (AFHMM) was used to separate appliances from the aggregated load data. The main motivation and contribution of this approach is that it addresses the local optima problems that existing approximate inference techniques~\cite{kim2011unsupervised} are highly susceptible to experience. The idea is to exploit the  additive structure of AFHMM to develop a convex formulation of approximate inference that is more computationally efficient and has no issues of local optima. Although, this approach was applied on  relatively high frequency electricity data~\cite{kolter2011redd}, the data scale is not close to ours. Hierarchical Dirichlet Process Hidden Semi-Markov Model (HDP-HSMM) is used in~\cite{johnson2013bayesian} to incorporate duration distributions (Semi Markov) and allows to infer the number of states from the data (Hierarchical Dirichlet Process). On the contrary, the AFHMM algorithm in~\cite{kolter2012approximate} requires the number of appliances (states) to be set a-priori.

The common feature of the approaches discussed so far is that the considered data sets are collected only from the electricity signals. In contrast, our data involves different utilities namely electricity, water and gas data as well some sensors measurements that provide contextual features. To the best of our knowledge, the only data that considers water and gas usage data is~\cite{makonin2013ampds,makonin2016electricity}. However, the sampling rate of this data is very low compared to ours. Authors in~\cite{wytock2014contextually} exploit the correlation between appliances and side information, in particular temperature, in a convex optimisation problem for energy disaggregation. This algorithm is applied on low sampling rate electricity data with contextual supervision in the form of temperature information.

To wrap up this section, three features distinguish our approach from existing ones. It bridges the gap between pattern recognition and NILM making it beneficial for a variety of different applications. Driven by massive amount of data, our method is computationally efficient and scalable, unlike state-of-the-art probabilistic methods
that posit detailed temporal relationships and involve complex inference steps. The available data has a high sampling rate electricity data allowing extracting more informative features and includes data from other utility usage and additional sensors measurements. Thus, our work also covers the research aspect of NILM concerned with the acquisition of data, prepossessing steps and evaluation of  NILM algorithms. 
\section{The Approach}\label{sec3}
In this section, we present the proposed approach which consists of two steps: features extraction and  pattern mining. First, a background on stochastic variational inference for a family of graphical models is provided. Next, we derive the pattern mining algorithm, online GLDA, which is an instance of the family of graphical models and operates online to accommodate high volume and speed data streams. Finally, we present the feature extraction step and summarise the full algorithm.

\subsection{Background: Stochastic Variational Inference}\label{subsec1}
In the following, we describe the model family of which GLDA is a member and review SVI.

\textbf{Model family.} The family of models considered here consists of three random variables: observations $\boldsymbol x=\boldsymbol{x}_{1:D}$, local hidden variables $\boldsymbol z=\boldsymbol{z}_{1:D}$, global  hidden variables $\boldsymbol\beta$ and fixed parameters $\boldsymbol\alpha$. The model assumes that the distribution of  the $D$ pairs of $(\boldsymbol{x}_i,\boldsymbol{z}_i)$ is conditionally independent given $\boldsymbol\beta$. Furthermore, their distribution and the prior distribution of $\boldsymbol\beta$ belong to the exponential family.
\begin{equation}\label{equ1}
p(\boldsymbol\beta,\boldsymbol x,\boldsymbol z|\boldsymbol\alpha)=p(\boldsymbol\beta|\boldsymbol\alpha)\prod_{i=1}^D p(\boldsymbol{z}_i,\boldsymbol{x}_i|\boldsymbol\beta)
\end{equation}
\begin{equation}\label{equ2}
p(\boldsymbol{z}_i,\boldsymbol{x}_i|\boldsymbol\beta)=h(\boldsymbol{x}_i,\boldsymbol{z}_i)\exp\big(\boldsymbol\beta^Tt(\boldsymbol x_i,\boldsymbol z_i)-a(\boldsymbol\beta)\big)
\end{equation}
\begin{equation}\label{equ3}
p(\boldsymbol\beta|\boldsymbol\alpha)=h(\boldsymbol\beta)\exp\big(\boldsymbol\alpha^Tt(\boldsymbol\beta)-a(\boldsymbol\alpha) \big)
\end{equation}
Here, we overload the notation for the base measures $h(.)$, sufficient statistics $t(.)$ and log normalizer $a(.)$. While the soul of the proposed  approach is generic, for simplicity we assume a conjugacy relationship between $(\boldsymbol x_i,\boldsymbol z_i)$ and $\boldsymbol\beta$. That is, the distribution $p(\boldsymbol\beta|\boldsymbol x,\boldsymbol z)$ is in the same family as the prior $p(\boldsymbol\beta|
\boldsymbol\alpha)$.

Note that this innocent looking family of models includes (but is not limited to) latent Dirichlet allocation \cite{blei2003latent}, Bayesian Gaussian mixture, probabilistic matrix factorization, hidden Markov models, hierarchical  linear and probit regression, and many Bayesian non-parametric models.  

\textbf{Mean-field variational inference.} Variational inference (VI) approximates intractable posterior $p(\boldsymbol\beta,\boldsymbol z|\boldsymbol x)$ by positing a family of simple distributions $q(\boldsymbol\beta,\boldsymbol z)$ and find the member of the family that is closest to the posterior (closeness is measured with KL divergence). The resulting optimization problem is equivalent maximizing the evidence lower bound (ELBO):
\begin{equation}\label{equ4}
\mathcal{L}(q)=E_q[\log p(\boldsymbol x,\boldsymbol z,\boldsymbol\beta)]-E_q[\log p(\boldsymbol z\boldsymbol\beta)]\leq \log p(\boldsymbol x)
\end{equation}
 Mean-field is the simplest family of distribution, where the distribution over the hidden variables factorizes as follows:
 \begin{equation}\label{equ5}
 q(\boldsymbol\beta,\boldsymbol z)=q(\boldsymbol\beta|\boldsymbol\lambda)\prod_{i=1}^Dp(\boldsymbol z_i|\boldsymbol\phi_i)
 \end{equation}
 Further, each variational distribution is assumed to come  from the same family of the true one. Mean-field variational inference optimizes the new ELBO with respect to the  local and global variational parameters $\boldsymbol\phi$ and $\boldsymbol\lambda$.
 \begin{equation}\label{equ6}
\mathcal{L}(\boldsymbol\lambda,\boldsymbol\phi)=E_q\bigg[\log\frac{p(\boldsymbol \beta)}{q(\boldsymbol \beta)}\bigg]+\sum_{i=1}^DE_q\bigg[\log\frac{p(\boldsymbol x_i,\boldsymbol z_i|\boldsymbol\beta)}{q(\boldsymbol z_i)} \bigg]
 \end{equation}
  It iteratively updates each variational parameter holding the other parameters fixed. With the assumptions taken so far, each update has  a closed form solution. The local parameters are a function of the global parameters.
 \begin{equation}\label{equ7}
 \boldsymbol\phi({\boldsymbol\lambda}_t)=\arg\max_{\boldsymbol\phi}\mathcal{L}(\boldsymbol\lambda_t,\boldsymbol\phi)
\end{equation}  
  We are interested in the global parameters which summarise the whole dataset (clusters in Bayesian Gaussian mixture, topics in LDA).
 \begin{equation}\label{equ8}
 \mathcal{L}(\boldsymbol\lambda)=\max_{\boldsymbol\phi} \mathcal{L}(\boldsymbol\lambda,\boldsymbol\phi)
\end{equation}  
To find the optimal value of $\boldsymbol\lambda$ given that $\boldsymbol\phi$ is fixed, we compute the natural gradient of $\mathcal{L}(\boldsymbol\lambda)$  and set it to zero by setting
 \begin{equation}\label{equ9}
\boldsymbol\lambda^*=\boldsymbol\alpha +\sum_{i=1}^DE_{\boldsymbol\phi_i({\boldsymbol\lambda}_t)}[t(\boldsymbol x_i,\boldsymbol z_i)]
 \end{equation} 
Thus, the new optimal global parameters are $\boldsymbol\lambda_{t+1}=\boldsymbol\lambda^*$.  The algorithm works by iterating between computing  the optimal local parameters  given the global ones \big(Eq.~\ref{equ7}\big) and computing the optimal global parameters given the local ones \big(Eq.~\ref{equ9}\big).
 
\textbf{Stochastic variational inference.} Rather than analysing  all the data to compute $\boldsymbol\lambda^*$ at each iteration, stochastic optimization can be used. Assuming that the data samples are uniformly randomly selected from the dataset, an unbiased noisy estimator of $\mathcal{L}(\boldsymbol\lambda,\boldsymbol\phi)$  can be developed based on a single data point. 
\begin{equation}\label{equ10}
\mathcal{L}_i(\boldsymbol\lambda,\boldsymbol\phi_i)=E_{q}\bigg[\log\frac{p(\boldsymbol \beta)}{q(\boldsymbol \beta)}\bigg]+DE_q\bigg[\log\frac{p(\boldsymbol x_i,\boldsymbol z_i|\boldsymbol\beta)}{q(\boldsymbol z_i)} \bigg]
\end{equation}\label{equ11}
The unbiased stochastic approximation of the ELBO as a function of $\boldsymbol\lambda$ can be written as follows
\begin{equation}\label{equ11b}
\mathcal{L}_i(\boldsymbol\lambda)=\max_{\boldsymbol\phi_i}\mathcal{L}_i(\boldsymbol\lambda,\boldsymbol\phi_i)
\end{equation}
Following the same steps in the previous section, we end up with a noisy unbiased estimate of Eq.~\ref{equ8} 
\begin{equation}\label{equ12}
\boldsymbol{\hat{\lambda}}=\boldsymbol\alpha +DE_{\boldsymbol\phi_i({\boldsymbol\lambda}_t)}[t(\boldsymbol x_i,\boldsymbol z_i)]
\end{equation}
At each iteration, we move the global parameters a step-size $\rho_t$ (learning rate) in the direction of the noisy natural gradient.
 \begin{equation}\label{equ13}
\boldsymbol\lambda_{t+1}=(1-\rho_t)\boldsymbol\lambda_t+\rho_t\boldsymbol{\hat{\lambda}}
\end{equation}
With certain conditions on $\rho_t$, the algorithm converges ($\sum_{t=1}^\infty\rho_t=\infty$, $\sum_{t=1}^\infty \rho_t^2<\infty$ )\cite{robbins1951stochastic}.

\subsection{Online Gaussian LDA}\label{subsec2}
Gaussian LDA (GLDA), as its name suggests, is an LDA with Gaussian components over the observations in place of the multinational ones of LDA. Hence, it is an instance of the family of models described in Sec~\ref{sec3} where the global, local, observed variables and their distributions are set as follows:
\begin{itemize}
\item the global variables $\{\boldsymbol\beta\}_{k=1}^K\equiv \{\boldsymbol\mu,\boldsymbol\Sigma\}_{k=1}^K$ are the components in GLDA. A component is a distribution over the input in the feature space, where the probability of an input vector $\boldsymbol x$ in component $k$, $p(\boldsymbol x|\boldsymbol\beta,k)=N(\boldsymbol x|\boldsymbol\mu_k,\boldsymbol\Sigma_k)$. Hence, the prior distribution of $\boldsymbol\beta_k$ is a Normal-Inverse-Wishart distribution $p(\boldsymbol\mu,\boldsymbol\Sigma)=\prod_kNIW(\boldsymbol\mu_k,\boldsymbol\Sigma_k|\boldsymbol m,\boldsymbol \omega,s,v)$.
\item The local variables are the component proportions $\{\boldsymbol\theta_d\}_{d=1}^D$ and the component assignments $\{\{z_{d,i}\}_{d=1}^D\}_{i=1}^{n}$ which index the Gaussian component that generate the observations. Each pattern is associated with a component proportion which is a distribution over components, $p(\boldsymbol\theta)=\prod_dDir(\boldsymbol\theta_d;\boldsymbol\alpha)$. The assignments  $\{\{z_{d,i}\}_{d=1}^D\}_{i=1}^{n}$  are indices, generated by $\boldsymbol\theta_d$, that couple components with observations, $p(\boldsymbol z_d|\boldsymbol\theta)=\prod_i\theta_{d,z_{d,i}}$.
\item The observations $\boldsymbol x_{d}$ are the observations during a specified period of time which are assumed to be drawn from components $\boldsymbol\beta$ selected by indices $\boldsymbol z_{d}$, $p(\boldsymbol x_d|\boldsymbol z_d,\boldsymbol\mu,\boldsymbol\Sigma)= \prod_{i}N(\boldsymbol x_{d,i}|\boldsymbol\mu_{z_{d,i}},\boldsymbol\Sigma_{z_{d,i}})$.
\end{itemize}
 The basic idea of GLDA is that each pattern is represented as random mixture over latent components, where each component is characterised by a distribution over the input observations. GLDA assumes the following generative process:
\begin{itemize}
\item[1]Draw components as follows: co-variance $\boldsymbol\Sigma_k\sim W^{-1}(\boldsymbol\omega,v)$; mean $\boldsymbol\mu_k\sim N(\boldsymbol m,\frac{1}{s}\boldsymbol\Sigma)$ for $k\in\{1,...,K\}$
\item[2]Draw component proportions $\boldsymbol\theta_d\sim Dir(\alpha,...,\alpha)$ for $d\in\{1,...,D\}$
\begin{itemize}
\item[2.1]Draw component assignments $z_{d,i}\sim Mult(\boldsymbol\theta_d)$ for $i\in\{1,...,n\}$
\begin{itemize}
\item[2.1.1] Draw an observation $x_{d,i}\sim N(\boldsymbol\mu_{z_{d,i}},\boldsymbol\Sigma_{z_{d,i}})$
\end{itemize}
\end{itemize}
\end{itemize}
According to Sec.~\ref{subsec1}, each variational distribution is assumed to come  from the same family of the true one. Hence, $q(\boldsymbol\beta_k|\boldsymbol\lambda_k)=NIW(\boldsymbol{qm}_k,\boldsymbol{q\omega}_k,qs_k,qv_k)$, $q(\boldsymbol\theta_d|\boldsymbol\gamma_d)=Dir(\boldsymbol\gamma_d)$ and $q(z_{d,i}|\boldsymbol\phi_{d,i})=Mult(\boldsymbol\phi_{d,i})$. To compute the update equations (shown in Eq.~\ref{equ13}) of the global variational parameters for GLDA, we need to find the sufficient statistic $t(.)$ presented in Eq.~\ref{equ2}. By writing the likelihood of GLDA in the form of Eq.~\ref{equ2}, we can obtain the following update:
\begin{align}\label{equ14}
\nonumber qs_k^{t+1}&=(1-\rho_t)qs_k^{t}+\rho_{t}(s+D\sum_{i=1}^n\phi_{d,i}^k)\\
\nonumber qv_k^{t+1}&=(1-\rho_t)qv_k^{t}+\rho_{t}(v+D\sum_{i=1}^n\phi_{d,i}^k)\\
\nonumber \boldsymbol{qm}_k^{t+1}&=(1-\rho_t)\frac{qs_k^{t}}{qs_k^{t+1}}\boldsymbol{qm}_k^{t}+\frac{\rho_t}{qs_k^{t+1}}(s\boldsymbol m+D\sum_{i=1}^{n}\phi_{d,i}^k\boldsymbol x_{d,i})\\
 \nonumber\boldsymbol{q\omega}_k^{t+1}&=(1-\rho_t)(\boldsymbol{q\omega}_k^{t}+qs_k^{t}\boldsymbol{qm}_k^{t}\boldsymbol{qm}_k^{tT})-qs_k^{t+1}\boldsymbol{qm}_k^{t+1}\boldsymbol{qm}_k^{t+1T}\\&+\rho_t(\boldsymbol\omega+s\boldsymbol m\boldsymbol m^T+D\sum_{i=1}^{n}\phi_{d,i}^k\boldsymbol x_{d,i}\boldsymbol x_{d,i}^T)
 \end{align}
where $d\sim \{1,...D\}$. The local variational parameters can be computed as follows:
\begin{align}\label{equ15}
\nonumber \phi_{d,i}^k&\propto exp\bigg(\Psi(\gamma_{d,i})-\frac{qv_k}{2}(\boldsymbol{x}_{d,i}-\boldsymbol{qm}_k)^T\boldsymbol{q\omega}_k^{-1}(\boldsymbol{x}_{d,i}-\boldsymbol{qm}_k)-\frac{F}{2qs_k}\\&\nonumber +\frac{1}{2}\big(\sum_{j=1}^{F}\Psi(\frac{qv_k+1-j}{2})+\log|\boldsymbol{q\omega}_k^{-1}|-F\log\pi\big)\bigg)\\
\boldsymbol\gamma_d&=\alpha+\sum_{i=1}^{n}\phi_{d,i}^k
\end{align}
where $F$ is the dimension of the feature space. Details on how Eq.~\ref{equ14} and \ref{equ15} are derived can be found in~\cite{hoffman2013stochastic,bishop2006pattern}. 

Having computed the main elements of the pattern recognition algorithm, we move to the next section in which the features extraction together with the pattern mining process are described.

\subsection{Features Extraction and Pattern Mining}\label{featuex}
The ultimate goal of the proposed approach is to provide a lower-dimensional representation expressing patterns in the data. To this end, the proposed approach consists of two steps: (1) features extraction and (2) pattern mining. In fact, features extraction is required to avoid the effects of the curse of dimensionality when applying the pattern mining algorithm (online GLDA). It helps reduce unnecessary redundancies in the raw data signal and extract informative features. Although, appliances identification is not the goal of this work, distinctive features providing useful information to discern appliances under use will be informative for GLDA. Such information suggests the activity performed by the residents leading to insight on their behaviour. In this work, features are extracted from the electricity signal only. Data coming from the other utilities and the sensors measurements are sampled at much lower rate, hence it is of small size. It feeds GLDA in their original representation.

Two main types of features have been proposed in the literature ~\cite{zoha2012non} with the purpose of detecting events: steady-state and transient event-based features. Steady state methods relate to changing operation state of the appliance; for example a change of steady-state active power measurement from a high to low value can identify whether the appliance is being turned On or Off. This kind of features can be captured with low sampling rate. The transient methods capture transient behaviour between steady-states; for example high frequency noise in electrical current or voltage, as a result of an appliance changing operation state. This type of features requires high sampling rate. Examples of features that can define appliance state transitions are shape, size, duration and the harmonics of the transient waveforms. These two types of features have been often used with supervised machine learning for appliances identification.

On the other hand, unsupervised algorithms like the ones discussed in Sec.~\ref{sec2} work directly  on separating  the power signal of individual appliance from the aggregated signal without preforming any sort of event detection. In contrast, our approach comprises the feature extraction phase, however, it is not an event detection method. We extract features that harness the high frequency of the sampling by exploiting information in the frequency spectrum as well as the conventional NILM features like the well-known reactive and active power features. Real and reactive power features have been shown to be very useful (alone or accompanied with other features) in many conventional non-intrusive load monitoring approaches~\cite{hart1992nonintrusive,marchiori2011circuit,norford1996non,marceau2000nonintrusive}. The importance of these features or features derived from them is that they convey information about the load of the appliance as well as the nature of it (difference between reactive and active power). Another advantage is that they do not require high sampling rate and therefore expensive current and voltage meters. However, the provided electricity data is sampled at high rate.

In addition, to exploit the information offered with the high sampling rate, we extract frequency domain features. We compute the RMS spectrum power over fixed bands of frequencies. The size and number of these bands are provided as parameter of the electricity extraction function. RMS spectrum power provides information about the waveforms. Different waveforms can characterise different types of appliances. Hence, it is expected that the frequency domain features will be very useful in the mining task. Adopting these features was also inspired by the work of the researchers at MIT as well as other research studies~\cite{lee2005estimation,laughman2003power,wichakool2009modeling,srinivasan2006neural,ruzzelli2010real}. The harmonics of the signal can also uniquely characterise non-linear loads that draw non-sinusoidal current during the operation. They have been used in combination with real and reactive power features~\cite{srinivasan2006neural,najmeddine2008state}. Detailed experiments for the features extraction are carried out in the next section.

These features are computed over windows of given duration granularity. Together with the gas water and other sensors data, they form a vector of observations. Specified number of these vectors are stacked over a pattern window to be used by the proposed algorithm (see fig.~\ref{fig1}). Algorithm~\ref{alg1} summarises the steps of the proposed method.

\begin{algorithm}[t]
   \caption{Pattern mining for energy consumption behaviour }\label{alg1}
\begin{algorithmic}[1]
       \STATE \textbf{Input:} raw-data window length, $R$; preprocessed-data window length, $n$; number of  preprocessed-data windows, $D$; number of GLDA's components, $K$; total number of iterations, $T$; learning rate,  $\{\rho_t\}_{t=0}^{T}$; GLDA's hyper-parameters, ($\boldsymbol m$,$\boldsymbol\omega$,$s$,$v$), $\alpha$.
       \STATE \textbf{Initialisation:} variational parameters: $\{(\boldsymbol {qm}_k^0, \boldsymbol{q\omega}_k^0, qs_k^0, qv_k^0)\}_{k=0}^{K}$.
       \FOR{$t=0,1,2,...T-1$}
        \STATE Read sequentially $n$ raw data windows of length $R$.
        \STATE Extract features  (see Sec.~\ref{featuex}) for each window
        \STATE Form windows of data points (actual input) of length $n$ in the new feature space ($\{\boldsymbol x_{d,i}\}_{i=1}^n$). 
        \STATE Initialise $\{\gamma_{dk}\}_{k=1}^K$
        \REPEAT
      \STATE Compute local variational parameters $\{\{\boldsymbol\phi_{d,i}^k\}_{i=1}^n\}_{k=1}^{K}$ (see Eq.~\ref{equ15})
      \STATE Update local variational parameters $\{\gamma_{dk}\}_{k=1}^K$ (see Eq.~\ref{equ15})
       \UNTIL{local parameters converge}
       \STATE Update global variational parameters $\{(\boldsymbol {qm}_k^0, \boldsymbol{q\omega}_k^0, qs_k^0, qv_k^0)\}_{k=0}^{K}$ (see Eq.~\ref{equ14})
     \ENDFOR
\end{algorithmic}
\end{algorithm}

\section{Empirical Evaluation}\label{sec4}

In this section we will first introduce the experimental data GLDA will be tested on along with details about the data pre-processing stages before results are provided.  

\subsection{Datasets}
The real-world multi-source utility usage data used here is provided by ETI\footnote{Energy Technologies Institute: http://www.eti.co.uk/}. The data includes electricity signals (voltage and current signals) sampled at high sampling rate around $205$ kHz, water and gas consumption sampled at low sampling rate. The data also contains other sensors measurements collected from the Home Energy Monitoring System (HEMS). 
 In this study we will use 4Tb of utility usage data collected from one  house over one month. This data has been recorded into three different formats. Water data is stored in text files with sampling rate of 10 seconds and is synchronised to Network Time Protocol (NTP) approximately once per month. Electricity data is stored in wave files with sampling rate of 4.88 μs and is synchronised to NTP every 28min 28sec. HEMS data is stored in a Mongo database with sampling rates differing according to the type of the data and sensors generating it (see Tab.~\ref{sensorsdatal}).
\begin{table}[t]
\centering
\caption{Characteristics of the data}
\label{sensorsdatal}
\begin{tabular}{|l|l|l|l|l|l|}
\hline
 Data&Range  &Resolution  &Measurement &Total \\ 
 &&  &frequency   &duration  \\\hline\hline
 Mains Voltage& -500V to +500V &62mV  &4.88µs  &1 months \\ \hline
 Mains Current& -10A to +10A &1.2mA  & 4.88µs    &1 months \\ \hline
 Water Flow Volume&0 to 100L per min  &52.4 pulses&10s&1 months\\
 &&per litre  & &    \\ \hline
 Room Air Temperature&0 to 40 DegC  &0.1 DegC  &Once every&1 months \\ &&&minute   &   \\ \hline
 Room Relative Humidity& 0 to 95\% & 0.1 \% &Once every&1 months\\
 &&&5 minutes  &    \\ \hline
 Hot Water&DegC&0.1 DegC&Once every&1 months\\
 Feed Temperature&  &  & 5 minutes  &    \\ \hline
 Boiler: Water&0 to 85 DegC&0.1 DegC&Once every&1 months\\ Temperature (Input)&  &  & 5 minutes  &    \\ \hline
 Boiler: Water&DegC&0.1 DegC&Once every&1 months \\ Temperature (Output)&  &  & 5 minutes  &   \\ \hline
Household: Mains Cold &DegC&0.1 DegC &Once every&1 months\\Water Inlet Temperature  &  & &  5 minutes &    \\ \hline
Gas Meter Reading   &Metric Meter  &0.01m3  &Once every&1 months\\
&&&15 minutes  &    \\ \hline
Radiator Temperature&DegC  &0.1 DegC  &Once every&1 months \\ &&&5 minutes   &   \\ \hline
Radiator Valve&0 to 100\% &50\%  &Once every&1 months \\ &&&5 minutes   &   \\ \hline
Boiler Firing Switch&Boolean &None  &Once every&1 months \\ &&&5 minutes   &   \\ \hline
\end{tabular}
\end{table}

\subsubsection{Data Pre-processing}
In order to exploit raw utility data by GLDA, a number of pre-processing steps are required. 
To do that, we implemented a Python code that reads the data from these different sources, synchronises its time-stamps to NTP time-stamps, extracts features and aligns the data samples to one time-stamp by measurement. For water data, the PC clock time-stamps of samples within each month are synchronised to NTP time-stamp. The synchronisation is done as follows:
\begin{equation}
 timestampNTP(i)=timestampsclock(i)+i\frac{Total\_Time\_Shift}{Number\_of\_Samples}   
\end{equation}

\begin{figure}[t]
\centering
\includegraphics[]{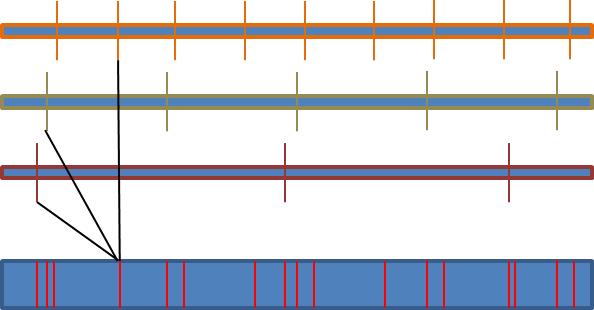}
\caption{Alignment of the data}
\label{fig2}
\end{figure}

In this equation, we assume that the total shift (between NTP and PC clock) can be distributed over the samples in one month. Similarly, Electricity data samples' time-stamps are synchronised to NTP time-stamps. The shift is distributed over 28 minutes and 28 seconds.
\begin{equation}
 timestampNTP(i)=timestampsclock(i)+i\frac{Total\_Time\_Shift}{Number\_of\_Samples}   
\end{equation}
The time-stamps of HEMS data were collected using NTP and so no synchronisation is required. Having all data samples synchronised to the same reference (NTP), we align the samples to the same time-stamps. The alignment strategy is shown in Fig.~\ref{fig2} where the union of all aligned data samples is stored in one matrix. Each row of this matrix includes a time-stamp and the corresponding values of the sensors. If for some sensors, there are no measurements taken at the time-stamp, the values measured at the previous time stamp are taken. 
The aligned data samples are the input of the feature extraction model. Pushed by the complexity of the mining task and motivated by the informativeness and simplicity of the water and sensors data, at this stage, we extract only few features from the electricity data over time windows of $1$ second (see Sec.~\ref{featuex}). These features are then aligned following the same process described earlier. Table~\ref{tab:features} shows the obtained features. 
\begin{table}[t]
\caption{Features after data pre-processing} 
\label{tab:features}
\centering
\scriptsize
\begin{tabular}{|c|c|p{0.8cm}|p{0.8cm}|p{0.8cm}|c|c|c|c|c|c|c|}\hline
\multirow{4}{*}{TimestampNTP} & \multirow{4}{*}{Water} &     \multicolumn{3}{c|}{Electricity} & \multicolumn{7}{c|}{HEMS} \\\cline{3-12}


&&\multirow{2}{*}{Real}&\multirow{2}{*}{React.}&\multirow{2}{*}{RMS}&\multirow{2}{*}{Gas}&\multicolumn{3}{c|}{Temperature}&\multirow{2}{*}{Humidity}&\multirow{2}{*}{Radiator}&\multirow{2}{*}{Boiler}\\\cline{7-9}

& & power&power &spectr. & & Rooms & Radiators & Water&&valve&firng\\\hline

$\vdots$ & $\vdots$ & $\vdots$ & $\vdots$ & $\vdots$ & $\vdots$ &
$\vdots$ & $\vdots$ & $\vdots$ & $\vdots$ & $\vdots$ & $\vdots$ \\ \hline

\end{tabular}
\end{table}




\subsection{Experimental Settings}

In all experiments, we use the empirical Bayes method to online point estimate the hyper-parameters from the data. The idea is to maximise the log likelihood of the data with respect to the hyper-parameters. Since the computation of the log likelihood of the data is not tractable, approximation based on the variational inference algorithm used in Sec.~\ref{subsec2} is employed. Following the same steps used to derive Eq.~\ref{equ14} and Eq.~\ref{equ15}, the update function for the hyper-parameters can be derived. The number of components is fixed to $K = 50$ 
We evaluated a range of settings of the learning parameters: $\kappa$, $\tau_0$ and batch size $BS$ on a validation set, where the parameters $\kappa$ and $\tau_0$, defined in~\cite{hoffman2010online}, control the learning step-size $\rho_t$. We used the data collected during the last week for validation and testing.

\subsection{Evaluation and Analysis}
In order to evaluate GLDA, we use the perplexity measure. Perplexity is used to quantify the fit of the model to the data. It is defined as the reciprocal geometric mean of the inverse marginal probability of the input in the held-out test set. Since perplexity cannot be computed directly, a lower bound on it is derived in a similar way to the one in~\cite{blei2003latent}. This bound is used as a proxy for the perplexity. 

Moreover, to investigate the quality of the results, we study the regularity of the mined patterns by matching them across similar periods of time. For instance, it is expected that similar patterns will emerge in specific hours like breakfast in every morning, watching TV in the evening, etc. Hence, it is interesting to understand how such patterns occur as regular events. 

Finally, to provide quantitative evaluation of the algorithm, we propose a mapping method that reveals the specific energy consumed for each pattern.  By doing so, we can evaluate numerically the coherence of the extracted patterns by fitting a regression model to the energy consumption over components:



\begin{equation}
A\boldsymbol{w}=\boldsymbol{b}  
\end{equation}
where $\boldsymbol{w}$ is a vector expressing energy consumption associated with components. $\boldsymbol b$ is a vector representing per-pattern consumption and $A$ is the matrix of the per-pattern components proportions obtained by GLDA. This technique will also allow  numerically checking the predicted consumption against the real consumption.

\subsubsection{A- Model Fitness:}
Although online GLDA converges for any valid $\kappa$, $\tau_0$ and $BS$, the quality and speed of the convergence may depend on how the learning parameters are set. We run online GLDA on the training sets for $\kappa\in \{0.1,0.3,0.6, 0.8, 0.9\}$, $\tau_0\in\{1, 64, 256, 1024\}$ and $BS\in \{1, 2, 4,5\}$. Table~\ref{my-label} summarises the best settings of each batch size along with the perplexity obtained on the test set.
\begin{table*}[t]
\centering
\caption{Parameter settings}
\label{my-label}
\begin{tabular}{|l|l|l|l|l|}
\hline
Batch size: $BS$& 1 & 2 & 4 &5   \\ \hline
Learning factor: $\kappa$    & 0.9 & 0.9 & 0.9& 0.9\\\hline
Learning delay: $\tau_0$    & 1024 & 1024 & 1024 & 1024  \\ \hline
Perplexity  &  4262700 &1027100  & 376558 & 332150 \\ \hline
\end{tabular}
\end{table*}    

The obtained results indicate that increasing the batch-size leads to better perplexity. However, the computation complexity
increases. Hence, we made a balance between model fitness and computation by setting the batch size to $4$, where the best learning parameters are $\kappa=0.9$ and $\tau=1024$. 

\subsubsection{B- Pattern Regularity:}
Using the optimal parameters' setting, we examine in the following the regularity of the mined patterns. To do that, we use the last two weeks of the data (from 18-05-2017 23:45:22 to 01-06-2017 23:45:22) for testing. To study the regularity of the energy consumption behaviour of the residents, we compare  the  mined patterns across different days of the testing period. These patterns are represented by the proportions of the different components (topics) inferred from the training data. To visualise the patterns, we plot gray-scale images showing the probability of different components with respect to the time. Black colour indicates probability of the component = 0, while white colour indicates probability = 1. Figure~\ref{f1} shows 14 figures split into two columns. The left column corresponds to the week from 18-05-2017 23:45:22 to 26-05-2017 23:45:22. The right column corresponds to the week from  26-05-2017 23:45:22 to 01-06-2017 23:45:22. Each figure depicts the pattern over 24 hours. The figures of the same days from different weeks are shown next to each other.  

\begin{figure}[p]
\centering
\subfigure{\includegraphics[width=0.49\textwidth]{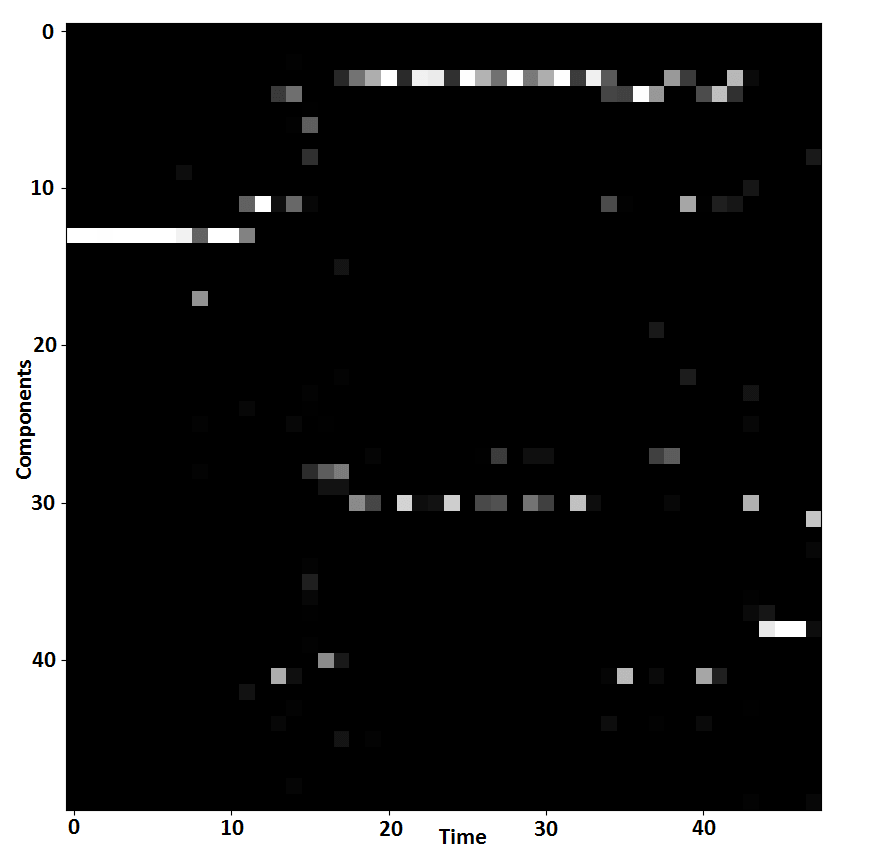}}
\subfigure{\includegraphics[width=0.49\textwidth]{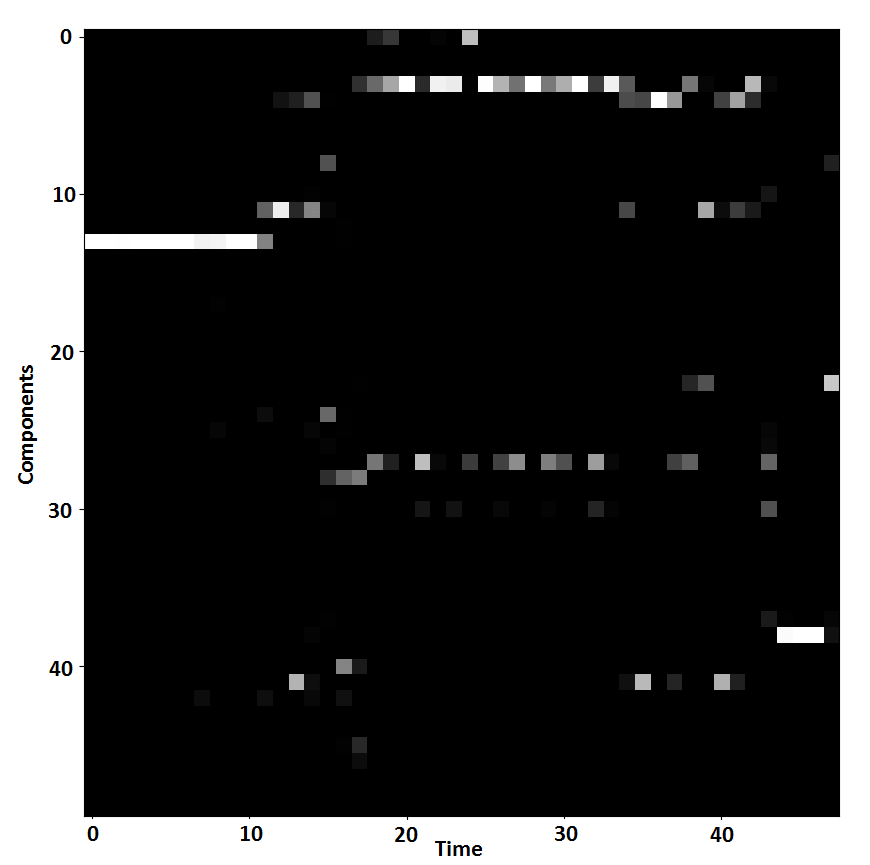}}\\
\subfigure{\includegraphics[width=0.49\textwidth]{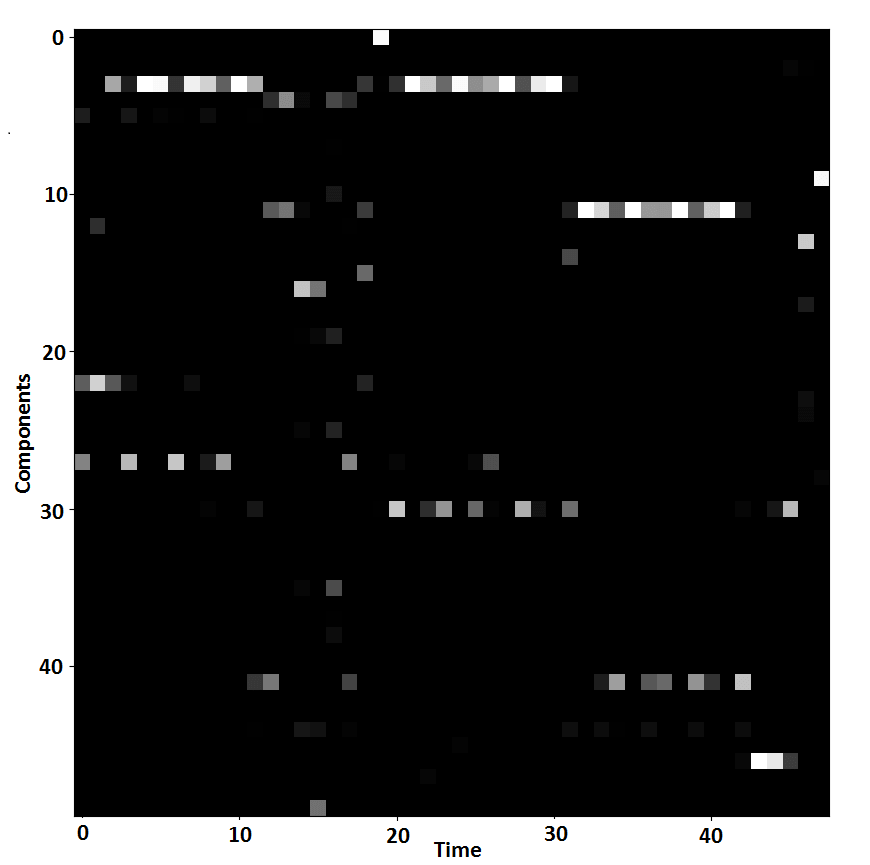}}
\subfigure{\includegraphics[width=0.49\textwidth]{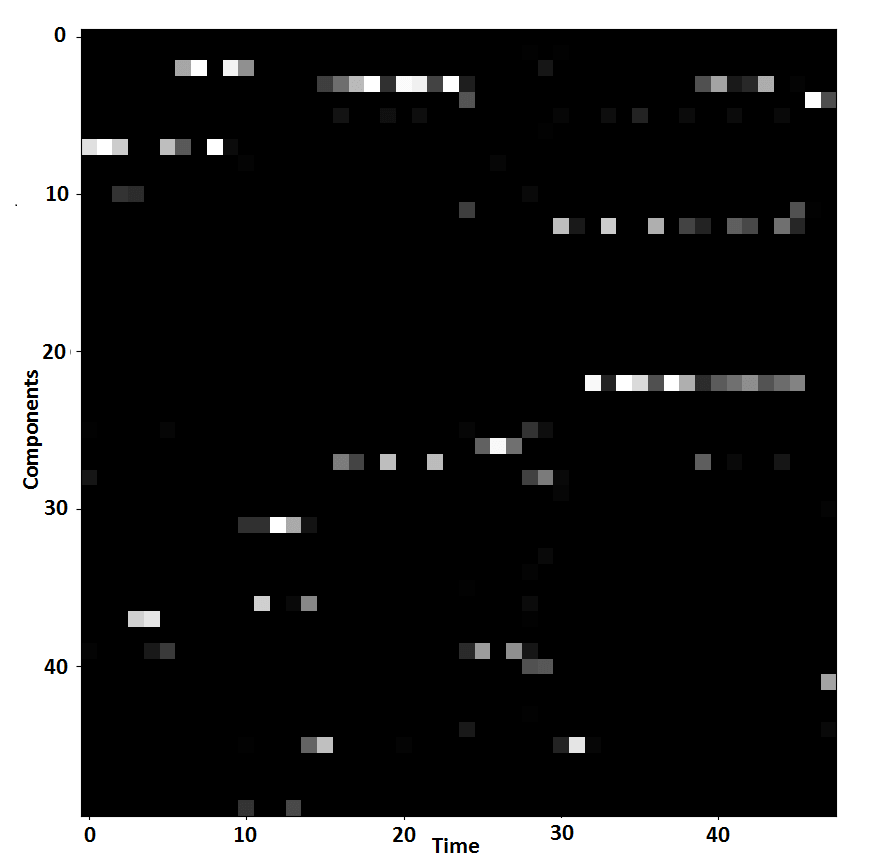}}\\
\subfigure{\includegraphics[width=0.49\textwidth]{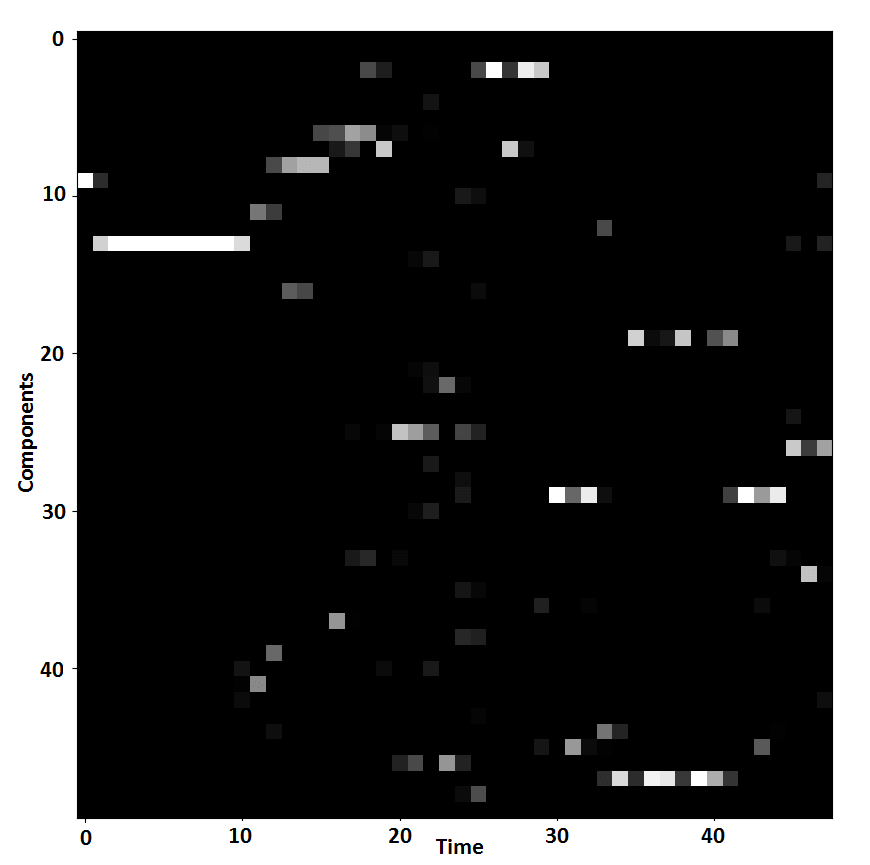}}
\subfigure{\includegraphics[width=0.49\textwidth]{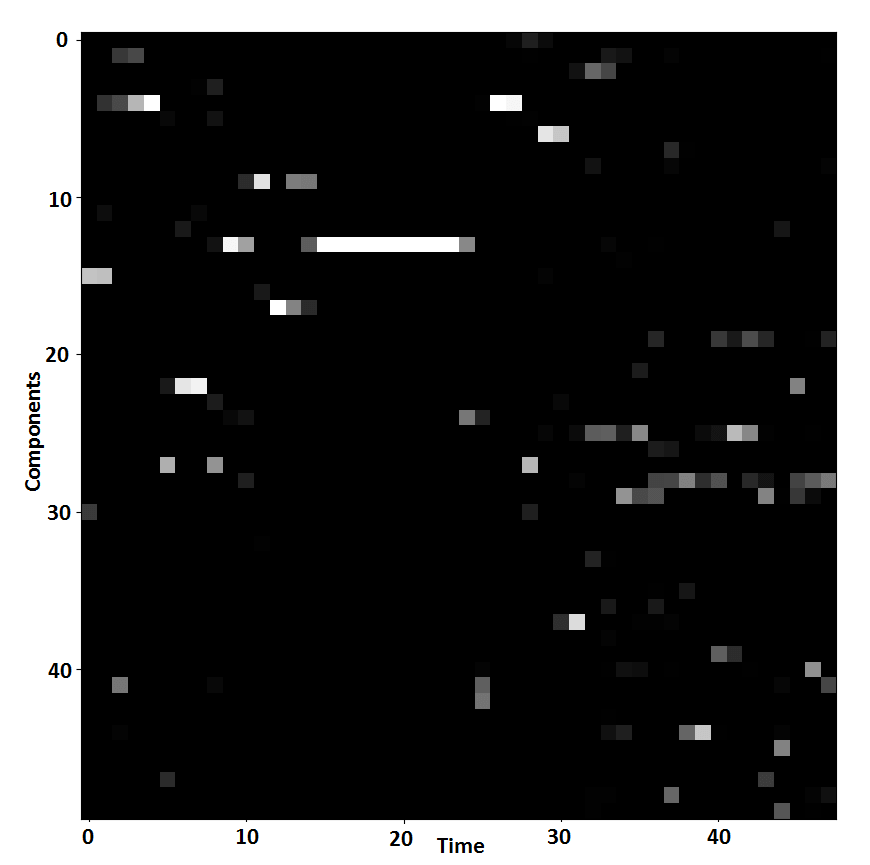}}\\
\subfigure{\includegraphics[width=0.49\textwidth]{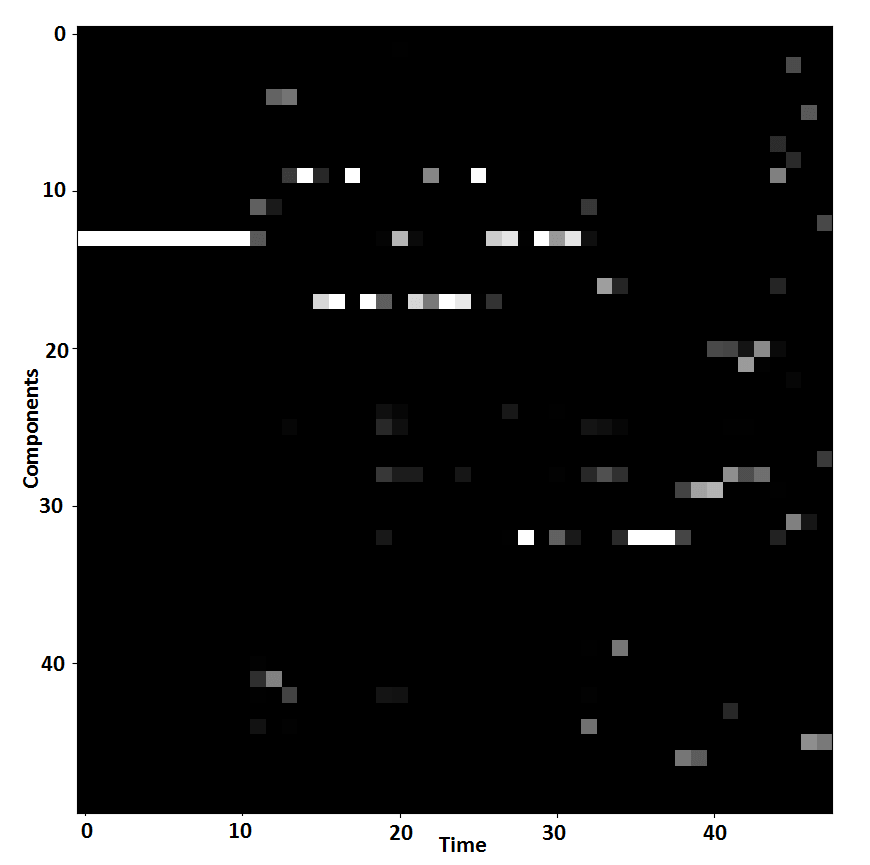}}
\subfigure{\includegraphics[width=0.49\textwidth]{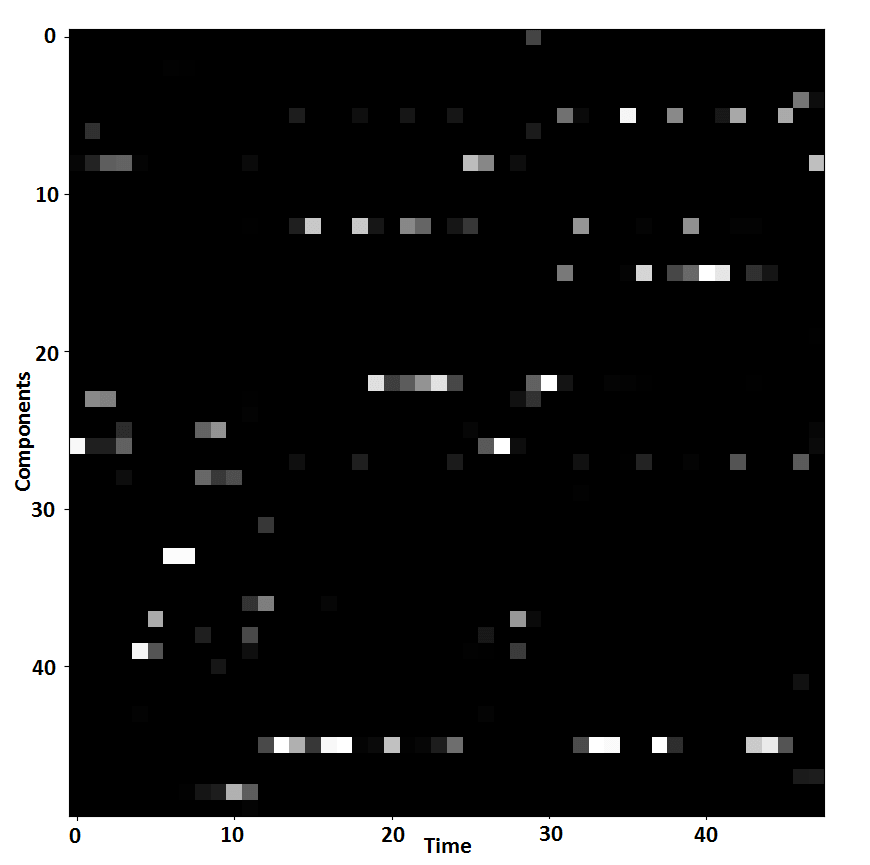}}\\
\end{figure} 
\begin{figure}[p]
\centering
\subfigure{\includegraphics[width=0.49\textwidth]{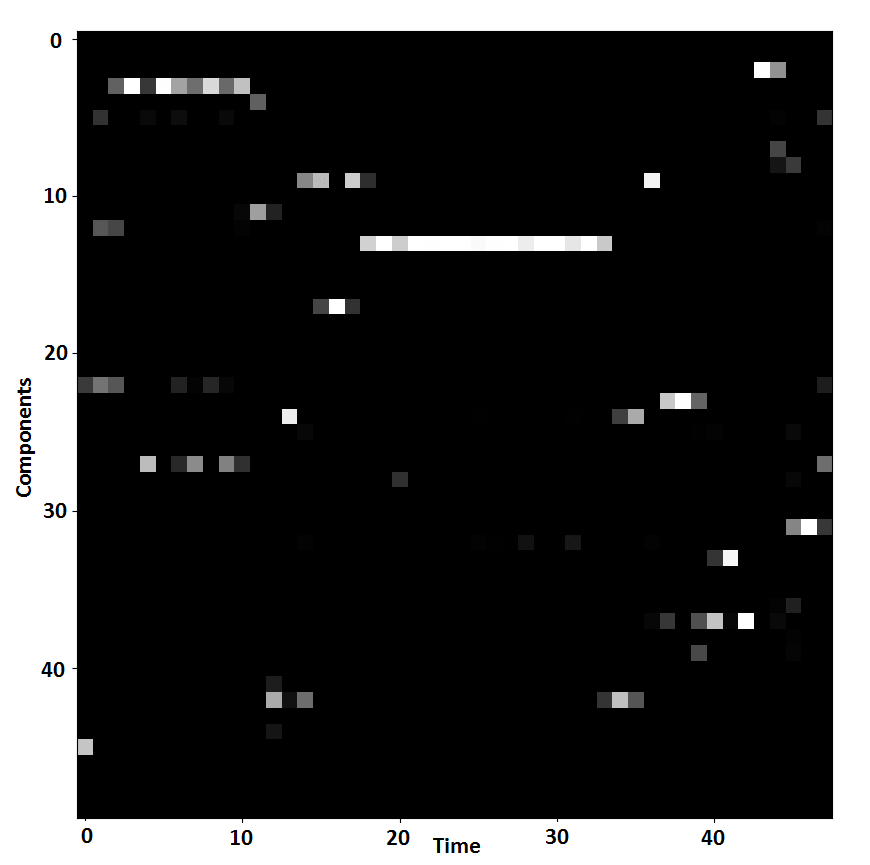}}
\subfigure{\includegraphics[width=0.49\textwidth]{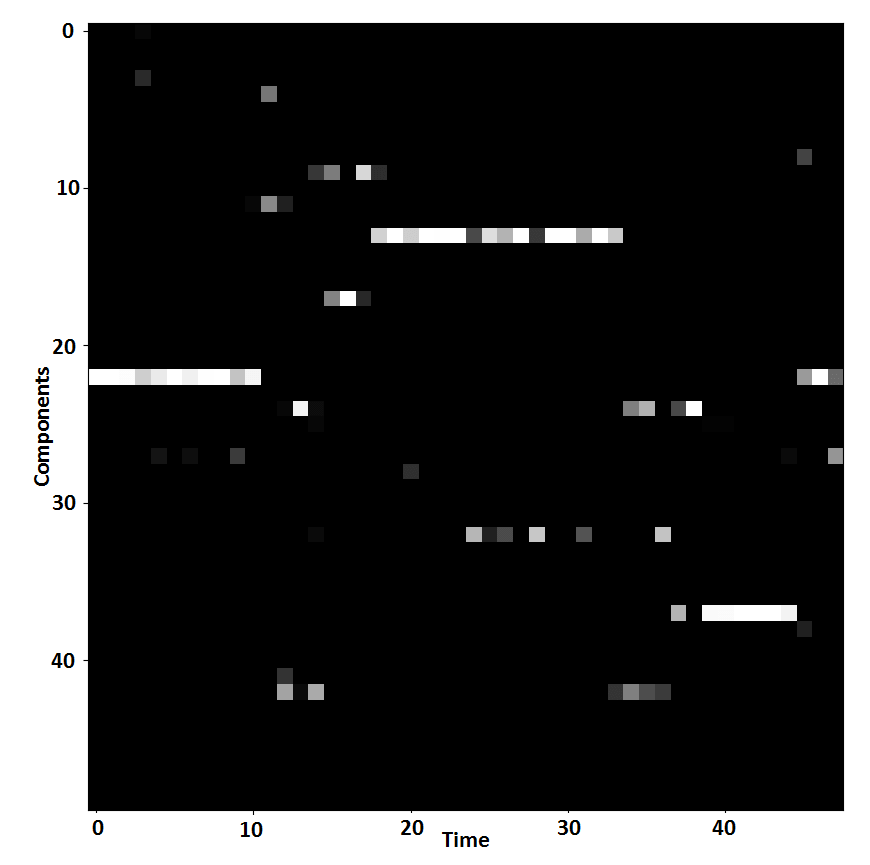}}\\
\subfigure{\includegraphics[width=0.49\textwidth]{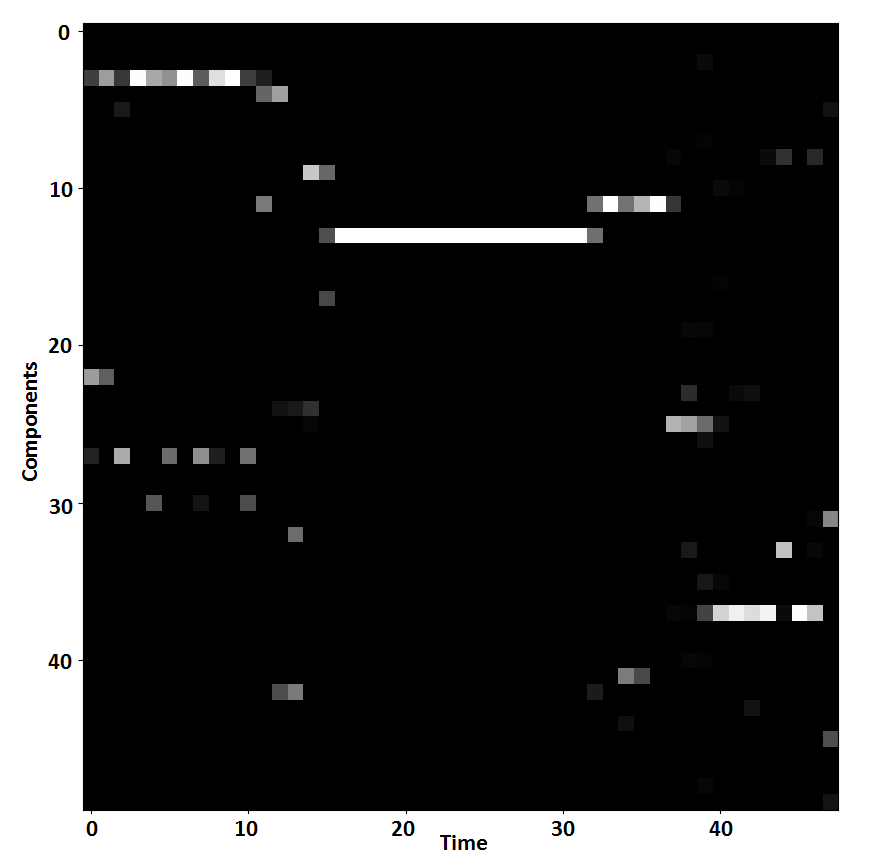}}
\subfigure{\includegraphics[width=0.49\textwidth]{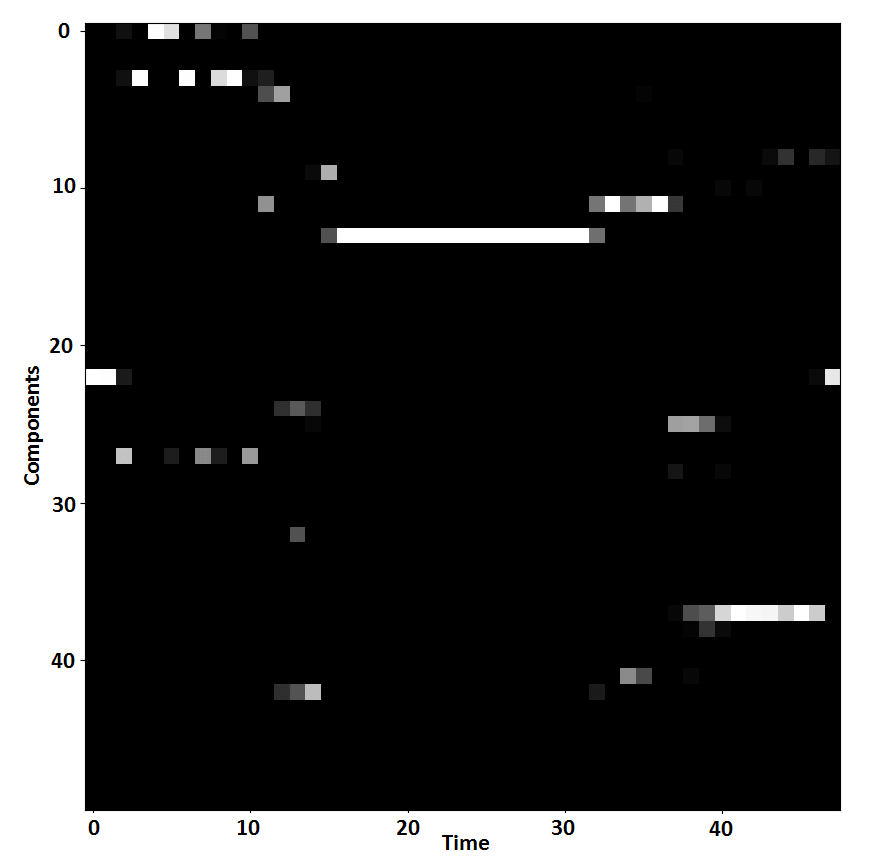}}\\
\subfigure{\includegraphics[width=0.49\textwidth]{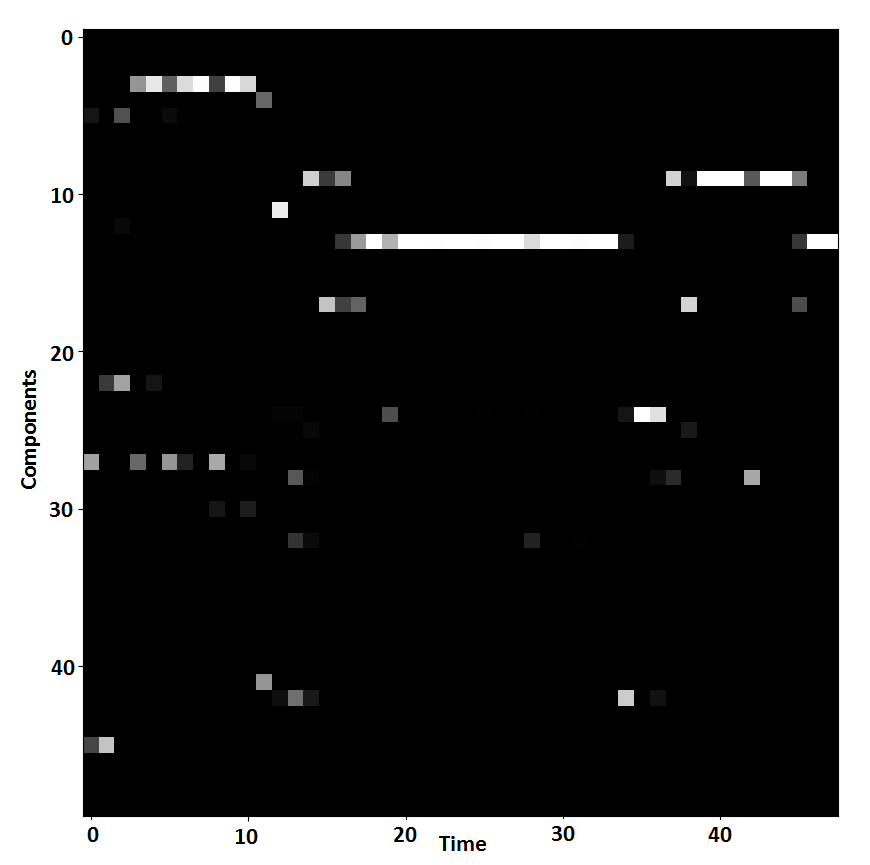}}
\subfigure{\includegraphics[width=0.49\textwidth]{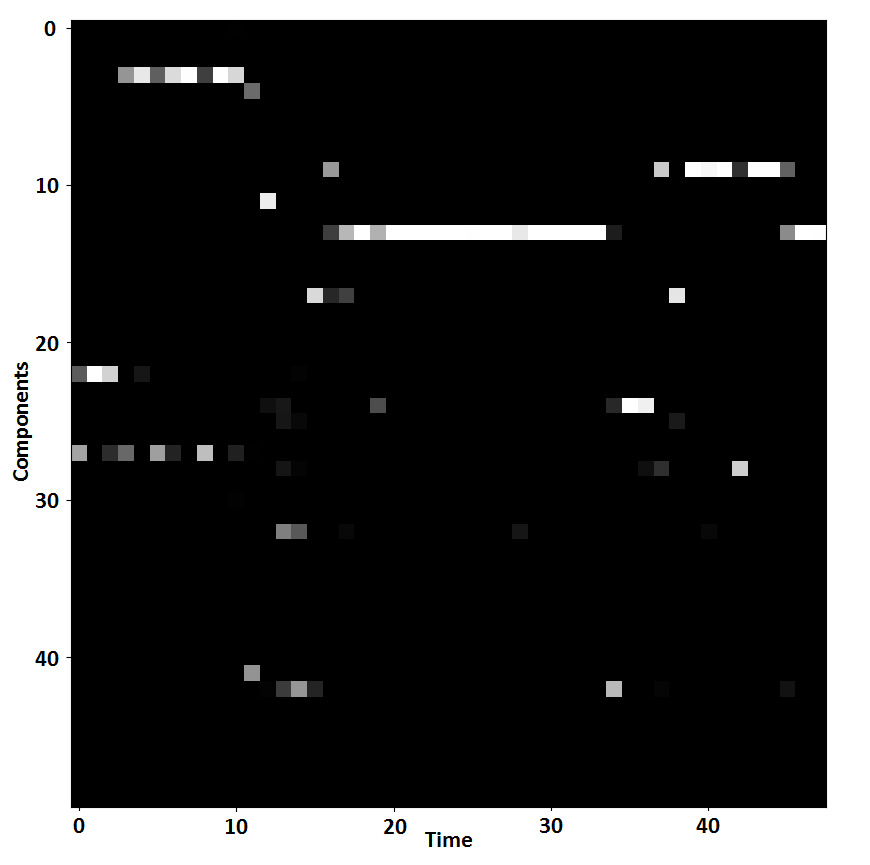}}\\
\caption{Emergence of patterns}\label{f1}
\end{figure} 

It can be clearly seen from these figures that there is regular patterns across columns. That is, similar energy consumption patterns appear across different weeks. Moreover, consumption patterns across working days within the same week are similar. On the other hand, for a specific week, the patterns over the weekend days and Friday are quite dissimilar to the rest within and across that week. This regularity may be caused by regular user lifestyle leading to similar energy consumption behaviour within and across the weeks. Such regularity is violated in the weekend, where more random activities could take place. Note that the difference between the patterns on 29-05-2017 (Monday) and that on 22-05-2017 (Monday) may be caused by the fact that on the 29th of May there was a bank holiday in the UK. Having shown that there is some regularity in the mined patterns, it is more likely that specific energy consumption can be associated with each component. In the next section, we apply a regression method to map the patterns (e.i., components proportions) to energy consumption. Thus, the parameters of interest are the energy consumption associated with the components. By attaching an energy consumption with each component, we can help validate the coherence of the extracted patterns and do forecasting. 

\subsubsection{C- Energy Mapping:}
\begin{figure}[!t]
\centering
\subfigure[Computed energy consumption]{\includegraphics[width=\textwidth]{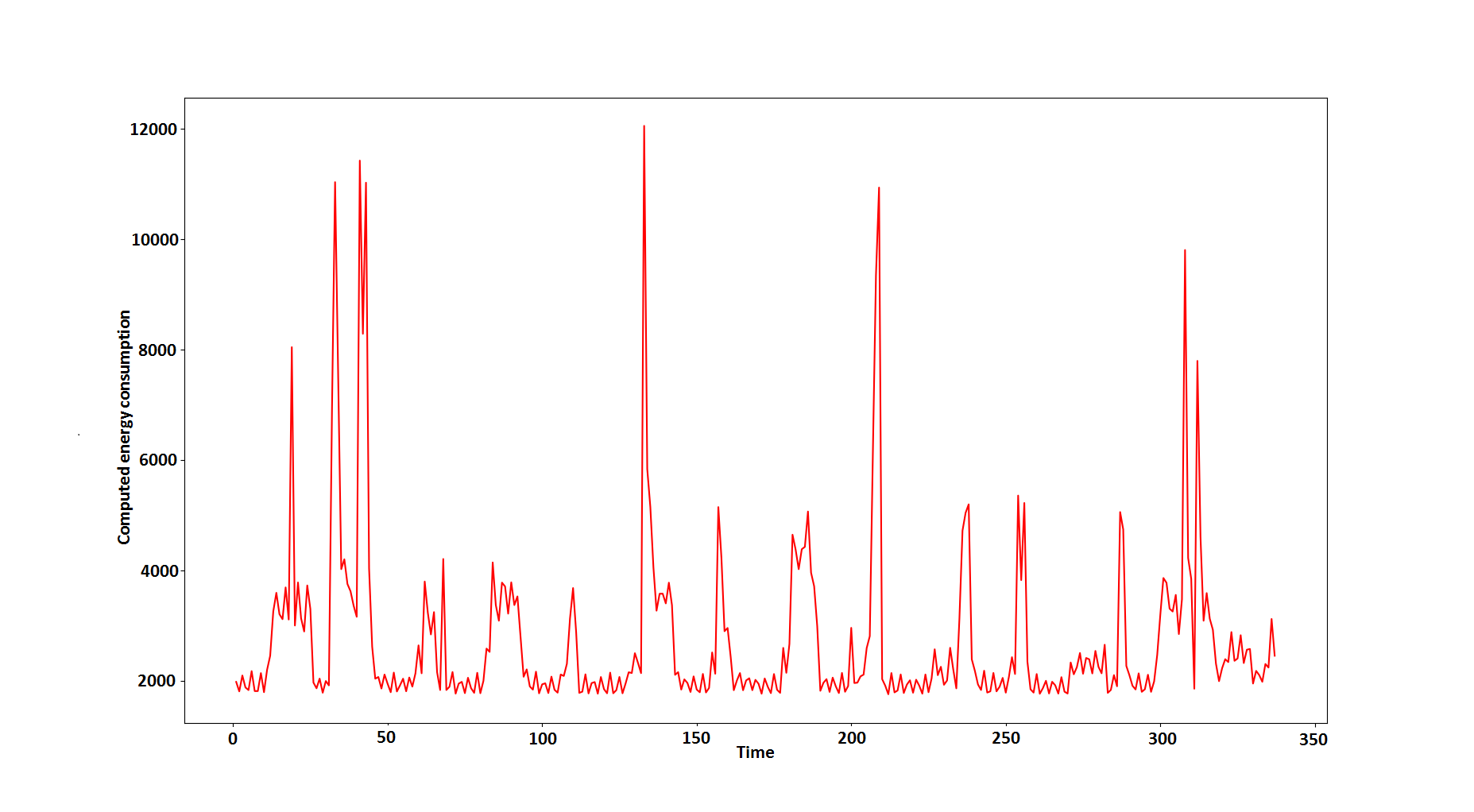}}\\
\subfigure[Estimated energy consumption]{\includegraphics[width=\textwidth]{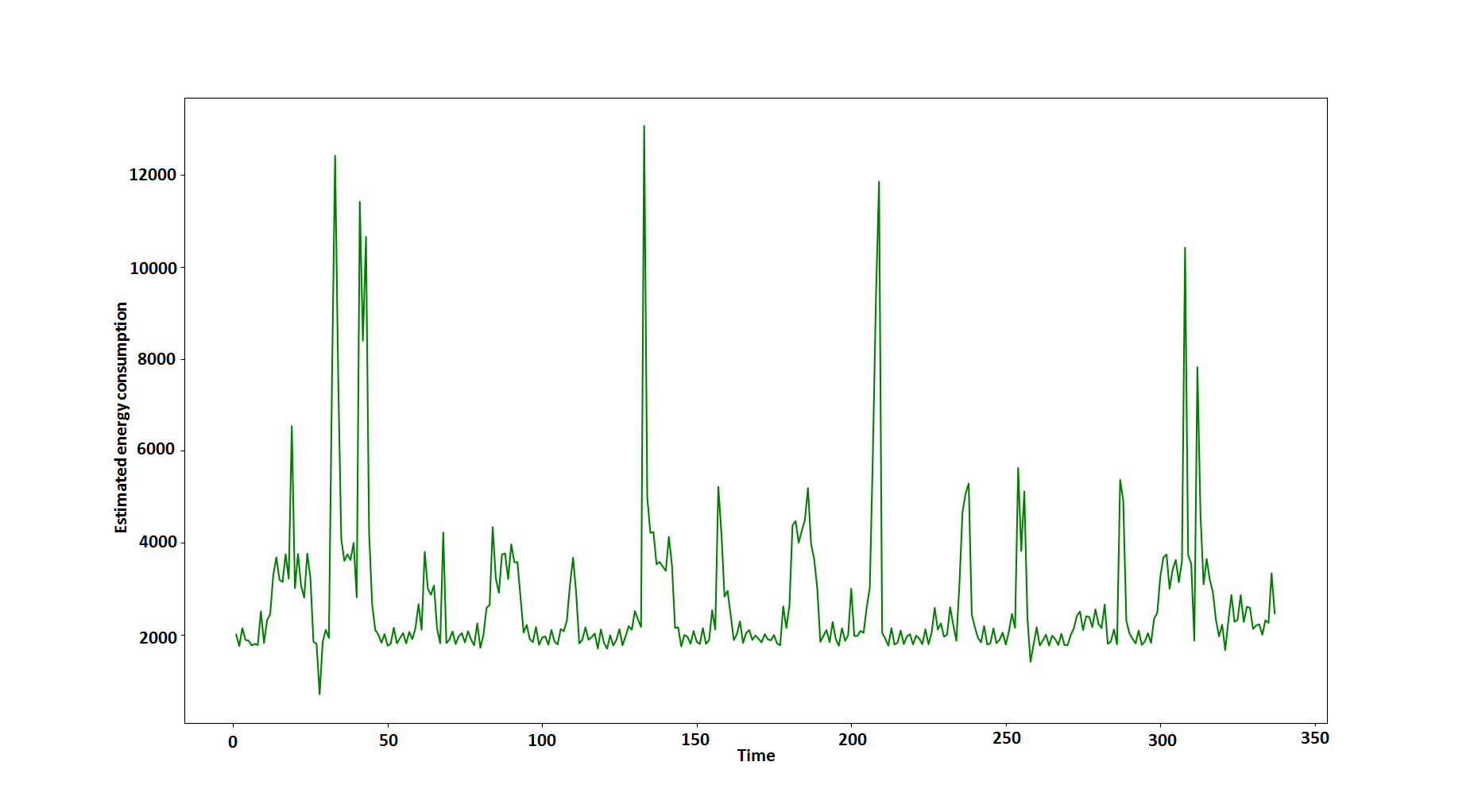}}
\caption{Evolution of the energy consumption over time }\label{f3}
\end{figure} 
As shown in the previous section, GLDA can express the energy consumption patterns by mixing  global components summarising data. These global components can be thought of as a base in the space of patterns. Each component is a distribution over a high-dimensional feature space and understanding what it represents is not easy. Hence, we propose to associate 
electricity 
consumption quantities to each component. Such association is motivated by the fact that an energy consumption pattern is normally governed by the usage of different appliances in the house. There should be a strong relation between components and appliances usage. Hence, a relation between components and energy consumption is plausible. Note that the best case scenario occurs if each component is associated with the usage of a specific appliance. Apart from the coherence study, associating energy consumption with each component can be used to forecast the energy consumption. This can be done through pattern forecasting which will be investigated in future work. More details will be given in Sec.~\ref{sec5}. We apply a simple least-square regression method to map patterns to energy consumption, expressed as follows:
\begin{equation}
\min ||A\boldsymbol{w}-\boldsymbol{b}||^2  
\end{equation}
where $\boldsymbol{w}$ is the per-component energy consumption vector, $\boldsymbol b$ is the per-pattern consumption vector and  $A$ is the matrix of the per-pattern components' proportions which is computed by GLDA. We train the regression model on the first testing week and run the model on the second one. Figure~\ref{f3} shows the energy consumption (in joules) along with the estimated consumption computed using the learned per-component consumption parameters.   

The similarity between the estimated and computed energy consumption demonstrates that the LDA components express distinct usages of energy. Such distinction can be the result of the usage of different appliances likely having distinct energy consumption signatures. Thus, the proposed approach produces coherent and regular patterns that reflect the energy consumption behaviour and human activities. Note that it is possible that different patterns (or appliance usages) may have the same energy consumption and that is why both estimated and computed energy consumption in Fig.~\ref{f3} are not fully the same.  
\section{Conclusion and Future Work}\label{sec5}
In this paper, we presented a novel approach, Gaussian LDA (GLDA), to extract patterns of the users' consumption behaviour from data involving different utilities (e.g, electricity, water and gas) as well as some sensors measurements. GLDA is fully unsupervised and works online which made it efficient for big data. To analyse the performance of GLDA, we proposed a three- step evaluation that covers: model fitness, qualitative analysis and quantitative analysis. The experiments show that the proposed method is capable of extracting regular and coherent patterns that highlight energy consumption over time. 

In the future, we foresee four directions for research to improve the obtained results and provide more features: (i) developing online dynamic Gaussian latent Dirichlet allocation (DGLDA) to consider the temporal dependency in the data leading to better results and allowing forecasting, (ii) replace the engineered features with ones extracted by a deep learning model trained directly on the raw data to yield richer low-level features, (iiii) develop more scalable GLDA by applying asynchronous distributed GLDA which can be derived from~\cite{mohamad2018asynchronous}  instead of SVI and (iiii) involve active learning strategy to query users about about ambiguous or unknown activities in order to guide the learning process when needed~\cite{mohamad2018active,mohamad2016bi}.
\section*{Acknowledgment}
This work was supported by the Energy Technology Institute (UK) as part of the project: \textit{High Frequency Appliance Disaggregation Analysis (HFADA)}.




\bibliographystyle{IEEEtran}
\bibliography{refer}

\begin{thebibliography}{10}
\providecommand{\url}[1]{#1}
\csname url@samestyle\endcsname
\providecommand{\newblock}{\relax}
\providecommand{\bibinfo}[2]{#2}
\providecommand{\BIBentrySTDinterwordspacing}{\spaceskip=0pt\relax}
\providecommand{\BIBentryALTinterwordstretchfactor}{4}
\providecommand{\BIBentryALTinterwordspacing}{\spaceskip=\fontdimen2\font plus
\BIBentryALTinterwordstretchfactor\fontdimen3\font minus
  \fontdimen4\font\relax}
\providecommand{\BIBforeignlanguage}[2]{{%
\expandafter\ifx\csname l@#1\endcsname\relax
\typeout{** WARNING: IEEEtran.bst: No hyphenation pattern has been}%
\typeout{** loaded for the language `#1'. Using the pattern for}%
\typeout{** the default language instead.}%
\else
\language=\csname l@#1\endcsname
\fi
#2}}
\providecommand{\BIBdecl}{\relax}
\BIBdecl

\bibitem{bulling2014tutorial}
A.~Bulling, U.~Blanke, and B.~Schiele, ``A tutorial on human activity
  recognition using body-worn inertial sensors,'' \emph{ACM Computing Surveys
  (CSUR)}, vol.~46, no.~3, p.~33, 2014.

\bibitem{poppe2010survey}
R.~Poppe, ``A survey on vision-based human action recognition,'' \emph{Image
  and vision computing}, vol.~28, no.~6, pp. 976--990, 2010.

\bibitem{wang2015review}
S.~Wang and G.~Zhou, ``A review on radio based activity recognition,''
  \emph{Digital Communications and Networks}, vol.~1, no.~1, pp. 20--29, 2015.

\bibitem{hart1992nonintrusive}
G.~W. Hart, ``Nonintrusive appliance load monitoring,'' \emph{Proceedings of
  the IEEE}, vol.~80, no.~12, pp. 1870--1891, 1992.

\bibitem{zeifman2011nonintrusive}
M.~Zeifman and K.~Roth, ``Nonintrusive appliance load monitoring: Review and
  outlook,'' \emph{IEEE transactions on Consumer Electronics}, vol.~57, no.~1,
  2011.

\bibitem{zoha2012non}
A.~Zoha, A.~Gluhak, M.~A. Imran, and S.~Rajasegarar, ``Non-intrusive load
  monitoring approaches for disaggregated energy sensing: A survey,''
  \emph{Sensors}, vol.~12, no.~12, pp. 16\,838--16\,866, 2012.

\bibitem{liang2010load}
J.~Liang, S.~K. Ng, G.~Kendall, and J.~W. Cheng, ``Load signature study—part
  i: Basic concept, structure, and methodology,'' \emph{IEEE transactions on
  power Delivery}, vol.~25, no.~2, pp. 551--560, 2010.

\bibitem{kolter2010energy}
J.~Z. Kolter, S.~Batra, and A.~Y. Ng, ``Energy disaggregation via
  discriminative sparse coding,'' in \emph{Advances in Neural Information
  Processing Systems}, 2010, pp. 1153--1161.

\bibitem{srinivasan2006neural}
D.~Srinivasan, W.~Ng, and A.~Liew, ``Neural-network-based signature recognition
  for harmonic source identification,'' \emph{IEEE Transactions on Power
  Delivery}, vol.~21, no.~1, pp. 398--405, 2006.

\bibitem{berges2009learning}
M.~Berges, E.~Goldman, H.~S. Matthews, and L.~Soibelman, ``Learning systems for
  electric consumption of buildings,'' in \emph{Computing in Civil Engineering
  (2009)}, 2009, pp. 1--10.

\bibitem{ruzzelli2010real}
A.~G. Ruzzelli, C.~Nicolas, A.~Schoofs, and G.~M. O'Hare, ``Real-time
  recognition and profiling of appliances through a single electricity
  sensor,'' in \emph{Sensor Mesh and Ad Hoc Communications and Networks
  (SECON), 2010 7th Annual IEEE Communications Society Conference on}.\hskip
  1em plus 0.5em minus 0.4em\relax IEEE, 2010, pp. 1--9.

\bibitem{kelly2015neural}
J.~Kelly and W.~Knottenbelt, ``Neural nilm: Deep neural networks applied to
  energy disaggregation,'' in \emph{Proceedings of the 2nd ACM International
  Conference on Embedded Systems for Energy-Efficient Built
  Environments}.\hskip 1em plus 0.5em minus 0.4em\relax ACM, 2015, pp. 55--64.

\bibitem{lai2013multi}
Y.-X. Lai, C.-F. Lai, Y.-M. Huang, and H.-C. Chao, ``Multi-appliance
  recognition system with hybrid svm/gmm classifier in ubiquitous smart home,''
  \emph{Information Sciences}, vol. 230, pp. 39--55, 2013.

\bibitem{bonfigli2015unsupervised}
R.~Bonfigli, S.~Squartini, M.~Fagiani, and F.~Piazza, ``Unsupervised algorithms
  for non-intrusive load monitoring: An up-to-date overview,'' in
  \emph{Environment and Electrical Engineering (EEEIC), 2015 IEEE 15th
  International Conference on}.\hskip 1em plus 0.5em minus 0.4em\relax IEEE,
  2015, pp. 1175--1180.

\bibitem{hoffman2013stochastic}
M.~D. Hoffman, D.~M. Blei, C.~Wang, and J.~Paisley, ``Stochastic variational
  inference,'' \emph{The Journal of Machine Learning Research}, vol.~14, no.~1,
  pp. 1303--1347, 2013.

\bibitem{fischer2008feedback}
C.~Fischer, ``Feedback on household electricity consumption: a tool for saving
  energy?'' \emph{Energy efficiency}, vol.~1, no.~1, pp. 79--104, 2008.

\bibitem{kelly2015uk}
J.~Kelly and W.~Knottenbelt, ``The uk-dale dataset, domestic appliance-level
  electricity demand and whole-house demand from five uk homes,''
  \emph{Scientific data}, vol.~2, p. 150007, 2015.

\bibitem{filip2011blued}
A.~Filip, ``Blued: A fully labeled public dataset for event-based non-intrusive
  load monitoring research,'' in \emph{2nd Workshop on Data Mining Applications
  in Sustainability (SustKDD)}, 2011, p. 2012.

\bibitem{kolter2011redd}
J.~Z. Kolter and M.~J. Johnson, ``Redd: A public data set for energy
  disaggregation research,'' 2011.

\bibitem{makonin2013ampds}
S.~Makonin, F.~Popowich, L.~Bartram, B.~Gill, and I.~V. Bajic, ``Ampds: A
  public dataset for load disaggregation and eco-feedback research,'' in
  \emph{Electrical Power \& Energy Conference (EPEC), 2013 IEEE}.\hskip 1em
  plus 0.5em minus 0.4em\relax IEEE, 2013, pp. 1--6.

\bibitem{makonin2016electricity}
S.~Makonin, B.~Ellert, I.~V. Baji{\'c}, and F.~Popowich, ``Electricity, water,
  and natural gas consumption of a residential house in canada from 2012 to
  2014,'' \emph{Scientific data}, vol.~3, 2016.

\bibitem{kim2011unsupervised}
H.~Kim, M.~Marwah, M.~Arlitt, G.~Lyon, and J.~Han, ``Unsupervised
  disaggregation of low frequency power measurements,'' in \emph{Proceedings of
  the 2011 SIAM International Conference on Data Mining}.\hskip 1em plus 0.5em
  minus 0.4em\relax SIAM, 2011, pp. 747--758.

\bibitem{kolter2012approximate}
J.~Z. Kolter and T.~Jaakkola, ``Approximate inference in additive factorial
  hmms with application to energy disaggregation,'' in \emph{Artificial
  Intelligence and Statistics}, 2012, pp. 1472--1482.

\bibitem{johnson2013bayesian}
M.~J. Johnson and A.~S. Willsky, ``Bayesian nonparametric hidden semi-markov
  models,'' \emph{Journal of Machine Learning Research}, vol.~14, no. Feb, pp.
  673--701, 2013.

\bibitem{wytock2014contextually}
M.~Wytock and J.~Z. Kolter, ``Contextually supervised source separation with
  application to energy disaggregation.'' 2014.

\bibitem{blei2003latent}
D.~M. Blei, A.~Y. Ng, and M.~I. Jordan, ``Latent dirichlet allocation,''
  \emph{Journal of machine Learning research}, vol.~3, no. Jan, pp. 993--1022,
  2003.

\bibitem{robbins1951stochastic}
H.~Robbins and S.~Monro, ``A stochastic approximation method,'' \emph{The
  annals of mathematical statistics}, pp. 400--407, 1951.

\bibitem{bishop2006pattern}
C.~M. Bishop, \emph{Pattern recognition and machine learning}.\hskip 1em plus
  0.5em minus 0.4em\relax springer, 2006.

\bibitem{marchiori2011circuit}
A.~Marchiori, D.~Hakkarinen, Q.~Han, and L.~Earle, ``Circuit-level load
  monitoring for household energy management,'' \emph{IEEE Pervasive
  Computing}, vol.~10, no.~1, pp. 40--48, 2011.

\bibitem{norford1996non}
L.~K. Norford and S.~B. Leeb, ``Non-intrusive electrical load monitoring in
  commercial buildings based on steady-state and transient load-detection
  algorithms,'' \emph{Energy and Buildings}, vol.~24, no.~1, pp. 51--64, 1996.

\bibitem{marceau2000nonintrusive}
M.~L. Marceau and R.~Zmeureanu, ``Nonintrusive load disaggregation computer
  program to estimate the energy consumption of major end uses in residential
  buildings,'' \emph{Energy conversion and management}, vol.~41, no.~13, pp.
  1389--1403, 2000.

\bibitem{lee2005estimation}
K.~D. Lee, S.~B. Leeb, L.~K. Norford, P.~R. Armstrong, J.~Holloway, and S.~R.
  Shaw, ``Estimation of variable-speed-drive power consumption from harmonic
  content,'' \emph{IEEE Transactions on Energy Conversion}, vol.~20, no.~3, pp.
  566--574, 2005.

\bibitem{laughman2003power}
C.~Laughman, K.~Lee, R.~Cox, S.~Shaw, S.~Leeb, L.~Norford, and P.~Armstrong,
  ``Power signature analysis,'' \emph{IEEE power and energy magazine}, vol.~99,
  no.~2, pp. 56--63, 2003.

\bibitem{wichakool2009modeling}
W.~Wichakool, A.-T. Avestruz, R.~W. Cox, and S.~B. Leeb, ``Modeling and
  estimating current harmonics of variable electronic loads,'' \emph{IEEE
  Transactions on power electronics}, vol.~24, no.~12, pp. 2803--2811, 2009.

\bibitem{najmeddine2008state}
H.~Najmeddine, K.~E.~K. Drissi, C.~Pasquier, C.~Faure, K.~Kerroum, A.~Diop,
  T.~Jouannet, and M.~Michou, ``State of art on load monitoring methods,'' in
  \emph{Power and Energy Conference, 2008. PECon 2008. IEEE 2nd
  International}.\hskip 1em plus 0.5em minus 0.4em\relax IEEE, 2008, pp.
  1256--1258.

\bibitem{hoffman2010online}
M.~Hoffman, F.~R. Bach, and D.~M. Blei, ``Online learning for latent dirichlet
  allocation,'' in \emph{advances in neural information processing systems},
  2010, pp. 856--864.

\bibitem{mohamad2018asynchronous}
S.~Mohamad, A.~Bouchachia, and M.~Sayed-Mouchaweh, ``Asynchronous stochastic
  variational inference,'' \emph{arXiv preprint arXiv:1801.04289}, 2018.

\bibitem{mohamad2018active}
S.~Mohamad, M.~Sayed-Mouchaweh, and A.~Bouchachia, ``Active learning for
  classifying data streams with unknown number of classes,'' \emph{Neural
  Networks}, vol.~98, pp. 1--15, 2018.

\bibitem{mohamad2016bi}
S.~Mohamad, A.~Bouchachia, and M.~Sayed-Mouchaweh, ``A bi-criteria active
  learning algorithm for dynamic data streams,'' \emph{IEEE transactions on
  neural networks and learning systems}, 2016.

\end{thebibliography}
\end{document}